\pdfoutput=1
\documentclass[11pt]{article}

\usepackage{todonotes}
\usepackage{changepage}
\usepackage{hyperref}
\usepackage{listings}

\usepackage{ACL2023}
\usepackage{ACL2023}
\usepackage{subfigure}
\usepackage{times}
\usepackage{latexsym}

\usepackage[T1]{fontenc}

\usepackage[utf8]{inputenc}

\usepackage{microtype}

\usepackage{inconsolata}

\usepackage{graphicx}
\usepackage{multirow}
\usepackage{comment}
\usepackage{enumitem}
\usepackage{listings}

\title{Polish-English medical knowledge transfer: A new benchmark and results}

    \author{
      Łukasz Grzybowski \\
    ARAAI, Poland \\
  Adam Mickiewicz University
  \And
  Jakub Pokrywka \\
  Adam Mickiewicz University\\
  \And
  Michal Ciesiółka \\
  Adam Mickiewicz University\\
  \AND
  Jeremi I. Kaczmarek \\
  Adam Mickiewicz University\\
  Poznan University of Medical Sciences \\
  \And
  Marek Kubis \\
  Adam Mickiewicz University\\
}

\begin{document}
\maketitle
\begin{abstract}

Large Language Models (LLMs) have demonstrated significant potential in specialized tasks, including medical problem-solving. However, most studies predominantly focus on English-language contexts. This study introduces a novel benchmark dataset based on Polish medical licensing and specialization exams (LEK, LDEK, PES). The dataset, sourced from publicly available materials provided by the Medical Examination Center and the Chief Medical Chamber, includes Polish medical exam questions, along with a subset of parallel Polish-English corpora professionally translated for foreign candidates. By structuring a benchmark from these exam questions, we evaluate state-of-the-art LLMs, spanning general-purpose, domain-specific, and Polish-specific models, and compare their performance with that of human medical students and doctors. Our analysis shows that while models like GPT-4o achieve near-human performance, challenges persist in cross-lingual translation and domain-specific understanding. These findings highlight disparities in model performance across languages and medical specialties, emphasizing the limitations and ethical considerations of deploying LLMs in clinical practice.
\end{abstract}

\section{Introduction}

The potential of Artificial Intelligence, especially Large Language Models (LLMs), is vast, but they come with considerable risks, particularly the issue of ``hallucinations'', where LLMs produce incorrect or misleading responses. This is especially concerning in fields like medicine, where errors can have serious consequences. Therefore, rigorous evaluation of LLM performance is essential before their clinical integration \cite{jeremi21}.

LLM performance varies significantly due to differences in training methods, datasets, and objectives, which affect their ability to perform specific tasks. The quality and diversity of training datasets are particularly important for specialized domains like medicine \cite{jeremi21}. While models trained on comprehensive, domain-specific datasets are expected to outperform those trained on general-purpose data, this assumption has been challenged 
\cite{nori2023can}.

Language also significantly impacts LLM performance. Most widely studied models are trained on multilingual datasets, predominantly in English, leading to better performance with English-language inputs and challenges with non-English content \cite{jeremi21}. Additionally, LLMs trained exclusively on non-English texts may lack important knowledge available only in English.

Modern medicine is evidence-based, and one might assume that the correct management of medical issues should be nearly universal. However, in practice, clinical practices are shaped by various factors, leading to significant variations in medical guidelines across countries. For instance, \citet{Zhou2024} analyzed 22 clinical practice guidelines from 15 countries, highlighting notable differences in recommendations for managing lower back pain.

LLMs trained primarily on English-language data are likely to align with disease prevalence and clinical guidelines typical of English-speaking countries. Consequently, their diagnostic and therapeutic recommendations may be biased towards practices common in these regions. When presented with the same clinical scenario in different languages, an LLM may produce varying responses, reflecting the diversity of healthcare practices across countries represented in the training data. Such discrepancies could be revealed by evaluating LLMs on non-English medical tests, like those conducted in Poland, where disease prevalence and medical guidelines may differ from those in English-speaking countries.

To primarily assess the performance of LLM models in medical question-answering tasks, we introduce a new benchmark based on publicly available exam questions from medical and dental licensing exams, as well as specialist-level exams conducted in Poland\footnote{The dataset is available at \url{https://huggingface.co/spaces/amu-cai/Polish_Medical_Exams}}. This dataset includes over 22,000 questions, primarily in Polish, with a subset of licensing exam questions also available in English, enabling comparative analysis. We propose a benchmark that enables the study of LLM behavior by addressing the following research questions: 
\setlist{nolistsep}
\begin{itemize}[noitemsep]
    \item How does the performance of LLMs on Polish medical examinations differ across various models and various exam types?
    \item How do LLMs compare to human doctors and medical students in performance? 
    \item How do LLMs' responses differ to general medical questions in Polish versus English, based on high-quality expert translations?
    \item What are the differences in the performance of LLMs on general versus specialized Polish medical exams?    
    \item How well do LLMs handle questions across various medical specialties (e.g., cardiology, neurology)? 
\end{itemize}


\section{Related work}

LLMs have the potential to revolutionize medicine by assisting medical professionals in key areas such as medical education, literature summarization, data extraction, manuscript drafting, and patient-clinical trial matching \cite{jeremi2, jeremi19}. They streamline communication by converting unstructured data to structured formats and simplifying documentation, such as summarizing patient records and generating medical reports \cite{jeremi16}. This reduces administrative burdens on clinicians, allowing more focus on patient care \cite{jeremi2}. LLMs also enhance personalized, patient-centered care, improve clinician-patient interactions, and may aid in diagnostics and management planning by analyzing medical data and monitoring patient parameters \cite{jeremi16, jeremi12}. 



Integrating LLMs into healthcare requires thorough evaluation to ensure reliability, safety, and equity, while identifying weaknesses and addressing biases to improve clinical care and support healthcare professionals \cite{jeremi8,jeremi9}. This evaluation should go beyond traditional performance metrics to include factors such as accuracy, reasoning, and factual reliability, using benchmarks like medical licensing exams, as well as assessing real-world utility, including clinical impact and workflow integration \cite{jeremi1}.

Small, fine-tuned BERT-style models continue to outperform LLMs in certain NLP tasks, such as text classification \cite{bucher2024fine}. However, the emergence of LLMs, such as GPT-3.5 and Med-PaLM 2, has led to significant advancements in medical question-answering benchmarks, including MedQA \cite{jin2021disease}, MedMCQA \cite{pal2022medmcqa}, and PubMedQA \cite{jin2019pubmedqa}. For this specific NLP task, general-purpose LLMs enhanced with specialized prompting strategies \cite{nori2023can} or fine-tuned domain-specific models surpass small encoder-only models in performance \cite{singhal2025toward}.


Most of the current datasets focus on English, which reflects both the dominance of English in medical research and the initial English-centric development of LLMs. However, there is growing recognition of the need for multilingual and non-English datasets to ensure the broader applicability of medical LLMs. MedQA is notable for its multilingual approach, incorporating questions from medical board exams in English, Simplified Chinese, and Traditional Chinese \cite{jin2021disease}. Additionally, there are datasets built around medical examinations in specific languages, including Swedish MedQA-SWE \cite{jeremi14}, Chinese CMExam \cite{jeremi4}, Japanese IGAKU QA \cite{jeremi10}, and Polish. 

For Polish, Lekarski Egzamin Końcowy (LEK, Eng. Medical Final Examination) is used as a benchmark \cite{jeremi11,jeremi7,jeremi5}. LEK is available in both Polish and English, allowing researchers to evaluate the influence of language on LLM performance. To date, analyses have primarily focused on GPT models, though several other LLMs, including LLaMa and Med42, have also been evaluated \cite{jeremi7}.

Regarding the Państwowy Egzamin Specjalizacyjny (PES, Eng. Polish Board Certification Examination), a few studies have assessed GPT's performance in specialized field exams \cite{jeremi5,jeremi18,wojcik}. \citet{jeremi20} provided a comprehensive evaluation of GPT-3.5 and GPT-4 on the PES, utilizing 297 exams across 57 specialties in Polish. 
We extend this research by incorporating additional PES specialties and introducing new exam types—the LEK and LDEK—in both Polish and English. Our study also includes cross-lingual evaluation of LLMs, comparisons with human performance, and assessments of publicly available LLMs, which were not done by \citet{jeremi20}.

\citet{10.1145/3589334.3645643} proposed a benchmark for the cross-lingual evaluation of LLMs. However, the questions they used were translated by a machine translation system, while the questions in our benchmark are translated by human medical experts from the examination center. Furthermore, we evaluated new models that demonstrate much better performance \cite{info15090543}.

\section{Polish medical exams dataset overview}

The LEK (Lekarski Egzamin Końcowy, Medical Final Examination) is a standardized exam for medical graduates and final-year students in Poland. Passing this exam, along with completing a postgraduate internship, is mandatory to obtain a medical license. Starting from 2022, 70\% of the questions come from a publicly available database, which includes 2,870 questions for LEK. The exam is conducted twice a year and lasts four hours, consisting of 200 multiple-choice questions. Candidates are allowed to retake the exam multiple times, even after passing, to improve their scores.

The LDEK (Lekarsko-Dentystyczny Egzamin Końcowy, Dental Final Examination) is the equivalent exam for dentistry graduates and final-year students, following the same format and requirements as the LEK.

The PES (Państwowy Egzamin Specjalizacyjny, National Specialization Examination) is a mandatory exam for physicians and dentists who have completed specialization training, including required internships and courses. It consists of a written test and an oral examination. The written test, held twice a year for each specialty, typically includes 120 multiple-choice questions, with one correct answer per question, and a passing score of 60\%. Candidates achieving at least 70\% on the written part are exempt from the oral examination, a rule introduced in late 2022. PES is considered the most challenging exam in the professional career of a medical doctor in Poland, and unlike LEK and LDEK, its questions are not made public before the exam.

In Poland, five types of exams for physicians and dentists are conducted: LEK (Lekarski Egzamin Końcowy, Eng. Medical Final Examination), LDEK (Lekarsko-Dentystyczny Egzamin Końcowy, Eng. Medical-Dental Final Examination), LEW (Lekarski Egzamin Weryfikacyjny, Eng. Medical Verification Examination), LDEW (Lekarsko-Dentystyczny Egzamin Weryfikacyjny, Eng. Medical-Dental Verification Examination), and PES (Państwowy Egzamin Specjalizacyjny, Eng. National Specialization Examination, Board Certification Exam). LEW and LDEW are for graduates of medical or dental studies carried outside of the European Union. Passing these exams is necessary for them to legally practice in Poland.\footnote{\url{https://www.cem.edu.pl/lew_info.php}\newline \url{https://www.cem.edu.pl/ldew_info.php}} However, these LEW and LDEW are taken by a relatively small number of candidates, and access to previous exam questions is limited. Therefore, they are not included in our work. The extensive descriptions of medical exams are included in Appendix \ref{appendix:detailedexams}.

The dataset comprises medical exams from the \href{https://cem.edu.pl/index.php}{Medical Examination Center} (Centrum Egzaminów Medycznych - CEM) and the \href{https://nil.org.pl/}{Supreme Medical Chamber} (Naczelna Izba Lekarska - NIL), covering LEK, LDEK, and PES exams from 2008–2024. It sources the exams as HTML quizzes and PDF files, with missing data from 2016–2020 (LEK/LDEK) and 2018–2022 (PES) partially filled using archives published on the NIL website. 
The exams are categorized by specialization, with questions and answers stored separately. Automated tools scrape and process the data, balancing parallelization with server constraints. Preprocessing ensures the dataset’s suitability for text-only AI benchmarks by removing irrelevant files, questions containing images, and content misaligned with current medical knowledge. We refer to these as "invalidated questions" throughout the text. Detailed descriptions of data sources, acquisition methods, and quality considerations appear in Appendix \ref{appendix:data_prep}.

\begin{table*}[h]
\centering
\small
\begin{tabular}{lrrrrr}
\hline
\textbf{Name} & \textbf{First} & \textbf{Last} & \textbf{Exams} & 
\textbf{Valid Questions} &  \textbf{Invalidated Questions}\\
\hline
LEK & 2008A & 2024S &  22 &4312 & 88 \\
LDEK & 2008A & 2024S & 22 & 4309 & 91\\
PES & 2008A & 2024S   &  72 & 8532 & 108\\
LEK (en)  & 2013A & 2024S & 14 & 2725 & 75 \\
LDEK (en)  & 2013A & 2024S & 14 & 2726 & 74\\
\hline
\textbf{total} & 2008A & 2024S & 144 & 22604 & 436 \\
\hline
\end{tabular}
\caption{Dataset statistics. S for Spring, A for Autumn.}
\label{tab:datasetstatistics}
\end{table*}


Finally, we create five sub-datasets: LEK, LDEK, PES, LEK en (LEK translated into English), and LDEK en (LDEK translated into English). Not all of them are released in the same edition, particularly the Polish and English counterparts. Therefore, the results presented in Section \ref{sec:perfrormance} should not be used to directly compare LLM performance on Polish exams with their English translations. To address this, we focus on the overlapping years and report these results in Section \ref{sec:crosslingual}. For the PES dataset, we collected a total of 180,712 questions. For the analysis in Sections \ref{sec:perfrormance}, \ref{sec:crosslingual}, and \ref{sec:humanresults}, we select only the most recent exam from each specialty and base our analysis on these exams. Detailed dataset statistics are provided in Table \ref{tab:datasetstatistics}, and example questions are presented in Appendix \ref{appendix:examplequestions}. In total, our analysis covers over 22,000 questions. For LLM inference, we use the Huggingface Transformers library \cite{transformers} and the OpenAI API.

\section{Performance of LLMs on exams}
\label{sec:perfrormance}

We categorize the models under study into the following groups: medical LLMs (models fine-tuned on English medical data), general-purpose multilingual LLMs, Polish-specific models, and models with restricted APIs.

\textbf{Medical Models:} \texttt{BioMistral-7B} \cite{labrak2024biomistral}, \texttt{Meditron-3} (8B and 70B versions) \cite{OpenMeditron}, \texttt{JSL-MedLlama-3-8B-v2.0} \cite{johnsnowlabsmedllama}.

\textbf{General-Purpose Multilingual Models:} \texttt{Qwen2.5 Instruct} (7B and 72B versions) \cite{qwen2.5}, \texttt{Llama-3.1 Instruct} (8B and 70B versions), \texttt{Llama-3.2-3B Instruct} \cite{dubey2024llama}, \texttt{mistralai/Mistral-Small-Instruct-2409}, and \texttt{Mistral-Large-Instruct-2407} \cite{mistral}.

\textbf{Polish-Specific Model:} \texttt{Bielik-11B-v2.2 Instruct} \cite{Bielik7Bv01}.

\textbf{Restricted API Models:} \texttt{GPT-4o-mini} and \texttt{GPT-4-o} \cite{openai2024gpt4technicalreport}.

We evaluate LLMs by directly prompting them to answer exam questions. Each prompt includes a brief introduction stating that the task is an exam for medical professionals consisting of single-choice questions. We do not provide additional examples or explanations in the prompt, and we do not use few-shot prompting. This approach aligns with the actual human exam environment, making it suitable for evaluating the models. Check \ref{appendix:prompts} for the exact prompts in Polish and English.\\
We report the models' results as the percentage of correct answers in Table \ref{tab:resultspercents} and the number of exams passed in Table \ref{tab:resultspassed}. Our findings are as follows: \texttt{GPT-4o} is the best performing model overall. 
Particularly in the PES category, \texttt{GPT-4o} outperforms the second-best model, \texttt{Meta-Llama-3.1-70B-Instruct}. \texttt{GPT-4o} is capable of passing all evaluated exams except for six PES exams. However, \texttt{GPT-4o-mini} performs significantly worse than \texttt{GPT-4o} and is also inferior to general-purpose open models. Among the open source models, \texttt{Meta-Llama-3.1-70B-Instruct} is the best performer. General-purpose models outperform medical-specific models, possibly because the latter were fine-tuned on English medical data.
The Polish-specific general-purpose model, \texttt{Bielik-11B-v2.2-Instruct}, performs worse than the top multilingual general-purpose models such as \texttt{Meta-Llama-3.1-70B-Instruct}, \texttt{Qwen2.5-72B-Instruct}, and \texttt{Mistral-Large-Instruct-2407}. However, for scenarios where deployment costs are more critical than performance, \texttt{Bielik-11B-v2.2-Instruct} may be preferable, as it still outperforms \texttt{Meta-Llama-3.1-8B-Instruct} of similar size in Polish-only exams.
Our final recommendation is to use \texttt{GPT-4o} for Polish medical data tasks. If using a restricted API is not feasible (e.g., due to patient anonymity requirements), \texttt{Meta-Llama-3.1-70B-Instruct} is suggested as an alternative.

\begin{table*}[h!]
\centering
\small
\begin{tabular}{lccccc}
\hline
\textbf{Model Name} & \textbf{LEK} & \textbf{LDEK} & \textbf{PES} & \textbf{LEK (en)} & \textbf{LDEK (en)} \\
\hline
BioMistral/BioMistral-7B & 25.86 & 24.58 & 23.32 & 32.92 & 26.71 \\
OpenMeditron/Meditron3-8B & 45.57 & 38.32 & 36.99 & 60.51 & 43.21 \\
OpenMeditron/Meditron3-70B & \textbf{66.93} & \textbf{47.20} & \textbf{47.42} & \textbf{67.05} & \textbf{45.71} \\
ProbeMedicalYonseiMAILab/medllama3-v20 & 40.61 & 34.05 & 31.79 & 52.40 & 38.15 \\
aaditya/Llama3-OpenBioLLM-70B & 55.15 & 39.78 & 40.06  & 66.09 & 45.27 \\
johnsnowlabs/JSL-MedLlama-3-8B-v2.0 & 36.46 & 31.17 & 28.89 & 54.13 & 39.40 \\
\hline
Qwen/Qwen2.5-7B-Instruct & 51.41 & 42.93 & 41.32 & 67.78 & 48.42 \\
Qwen/Qwen2.5-72B-Instruct & 76.39 & 59.50 & 59.14 & 82.24 & \textbf{62.95} \\
meta-llama/Meta-Llama-3.1-8B-Instruct & 51.02 & 42.38 & 39.91 & 65.03 & 47.40 \\
meta-llama/Meta-Llama-3.1-70B-Instruct & \textbf{80.47} & \textbf{63.40} & \textbf{61.71} & \textbf{83.01} & 62.73 \\
meta-llama/Meta-Llama-3.2-3B-Instruct & 39.31 & 33.77 & 32.69 & 52.59 & 37.09 \\
mistralai/Mistral-Small-Instruct-2409 & 51.37 & 40.98 & 38.35 & 64.04 & 43.03 \\
mistralai/Mistral-Large-Instruct-2407 & 76.32 & 58.71 & 59.52 & 82.61 & 61.85 \\
\hline
speakleash/Bielik-11B-v2.2-Instruct & \textbf{61.87} & \textbf{45.51} & \textbf{42.02} & \textbf{57.25} & \textbf{42.85} \\
\hline
gpt-4o-mini-2024-07-18 & 75.44 & 56.81 & 54.96 & 75.93 & 56.46 \\
gpt-4o-2024-08-06 & \textbf{89.40} & \textbf{75.63} & \textbf{75.35}  & \textbf{88.77} & \textbf{72.49} \\
\hline
\end{tabular}
\caption{The LLM results are represented as a percentage of correct answers of all datasets. The English versions of the LEK and LDEK exams are translated from the Polish versions; however, they represent only a subset of all the Polish exams.}
\label{tab:resultspercents}
\end{table*}

\begin{table*}[h!]
\centering
\small
\begin{tabular}{lrrrrrr}
\hline
\textbf{Model Name} & \textbf{LEK} & \textbf{LDEK} & \textbf{PES} & \textbf{LEK (en)} & \textbf{LDEK (en)} \\
\hline
BioMistral-BioMistral-7B & 0/22 & 0/22 & 0/72 & 0/14 & 0/14 \\
OpenMeditron-Meditron3-8B & 0/22 & 0/22 & 0/72 & \textbf{14/14} & 0/14 \\
OpenMeditron-Meditron3-70B & \textbf{22/22} & 0/22 & \textbf{7/72} & \textbf{14/14} & 0/14 \\
ProbeMedicalYonseiMAILab-medllama3-v20 & 0/22 & 0/22 & 0/72 & 3/14 & 0/14 \\
aaditya-Llama3-OpenBioLLM-70B & 16/22 & 0/22 & 0/72 & \textbf{14/14} & 0/14 \\
johnsnowlabs-JSL-MedLlama-3-8B-v2.0 & 0/22 & 0/22 & 0/72 & 4/14 & 0/14 \\
\hline
Qwen-Qwen2.5-7B-Instruct & 3/22 & 0/22 & 2/72 & \textbf{14/14} & 0/14 \\
Qwen-Qwen2.5-72B-Instruct & \textbf{22/22} & 19/22 & 32/72 & \textbf{14/14} & \textbf{14/14} \\
meta-llama-Meta-Llama-3.1-8B-Instruct & 2/22 & 0/22 & 1/72 & \textbf{14/14} & 0/14 \\
meta-llama-Meta-Llama-3.1-70B-Instruct & \textbf{22/22} & \textbf{21/22} & \textbf{46/72} & \textbf{14/14} & \textbf{14/14} \\
meta-llama-Llama-3.2-3B-Instruct & 0/22 & 0/22 & 0/72 & 3/14 & 0/14 \\
mistralai-Mistral-Small-Instruct-2409 & 2/22 & 0/22 & 0/72 & \textbf{14/14} & 0/14 \\
mistralai-Mistral-Large-Instruct-2407 & \textbf{22/22} & 16/22 & 30/72 & \textbf{14/14} & \textbf{14/14} \\
\hline
speakleash-Bielik-11B-v2.2-Instruct & \textbf{22/22} & \textbf{1/22} & \textbf{1/72} & \textbf{9/14} & \textbf{0/14} \\
\hline
gpt-4o-mini-2024-07-18 & \textbf{22/22} & 11/22 & 20/72 & \textbf{14/14} & 9/14 \\
gpt-4o-2024-08-06 & \textbf{22/22} & \textbf{22/22} & \textbf{68/72} & \textbf{14/14} & \textbf{14/14} \\
\hline
\end{tabular}
\caption{The LLM results are represented as a percentage of correct answers of all datasets. The LEK and LDEK exams are considered passed with a minimum score of 56\%, while the PES exam is considered passed with a minimum score of 60\%.}
\label{tab:resultspassed}
\end{table*}

The performance of LLMs varies significantly based on specialization in PES exams, which was noted by \cite{jeremi20} before. We provide a detailed analysis across specialties in Appendix \ref{appendix:specialities}, expanding upon the previous authors' findings with LLM other than the GPT family.


\section{Cross-lingual knowledge transfer}
\label{sec:crosslingual}

To compare the performance of various LLMs on Polish and English versions of the same datasets, we restrict the LEK and LDEK datasets to identical subsets. The English questions are translations of the original Polish questions, provided by human experts from the Medical Examination Center. Both versions are equivalent, meaning they convey the same medical content, structure, and intent, ensuring that the translated questions accurately reflect the original ones without altering their meaning or complexity. The analysis results, similar to the previous one, are presented in Tables \ref{tab:resultsenpercents} and \ref{tab:resultsenspassed}. As shown, all medical models, except for \texttt{OpenMeditron/Meditron3-70B}, perform better on the English versions of the datasets. This may be due to these models being fine-tuned on English medical corpora. General-purpose multilingual models perform better on the English versions of the exams as well. This result is anticipated since these models are trained on corpora containing significantly more English than Polish. While these models are proficient in Polish, their performance on the tests remains lower in Polish than in English. The difference can be considerable; for example, \texttt{meta-llama-Meta-Llama-3.1-8B-Instruct} passed only one LEK exam in Polish but passed all 13 when translated into English. However, as model quality improves, the performance gap between languages narrows. For instance, with \texttt{meta-llama-Meta-Llama-3.1-8B-Instruct}, the accuracy difference between Polish LEK (51.25\%) and English LEK (64.69\%) is 13.44 percentage points (or a 26\% relative change). In contrast, with \texttt{meta-llama-Meta-Llama-3.1-70B-Instruct}, the difference is only 1.66 percentage points (80.94\% for Polish LEK vs. 82.60\% for English LEK, or a 2\% relative change).

For \texttt{GPT-4o-mini}, which generally performs well, the results in English are only slightly better than in Polish. Interestingly, for \texttt{GPT-4o}, performance is actually higher on the Polish version. The only Polish LLM, Bielik, performs better on Polish LEK and slightly better on Polish LDEK, likely due to its fine-tuning from the multilingual model \texttt{Mistral-7B-v0.2} specifically for Polish. This fine-tuning enables it to better capture the nuances of Polish text to other models with a similar number of parameters. However, the tested \texttt{Bielik-v2.2-Instruct}, with only 11B parameters, is outperformed by models with double or even larger parameter counts on Polish versions of the LEK and LDEK exams. The only exceptions to this trend are \texttt{Mistral-Small-Instruct-2409} and \texttt{Llama3-OpenBioLLM-70B}.

Overall, our observations suggest that language transfer is more effective as the model’s general performance improves. Refer to Appendix \ref{appendix:detailed-cross-lingual} for detailed question-level analysis.

\begin{table*}[h!]
\centering
\small
\begin{tabular}{l|cc|cc}
\hline
\textbf{Model Name} & \textbf{LEK} & \textbf{LEK (en)} & \textbf{LDEK} & \textbf{LDEK (en)} \\
\hline
BioMistral/BioMistral-7B & 26.26 & \textbf{32.74} & 24.96 & \textbf{26.78} \\
OpenMeditron/Meditron3-8B & 45.99 & \textbf{60.34} & 37.97 & \textbf{43.35} \\
OpenMeditron/Meditron3-70B & \textbf{68.37} & 66.75 & \textbf{47.43} & 45.97 \\
ProbeMedicalYonseiMAILab/medllama3-v20 & 40.93 & \textbf{52.27} & 35.09 & \textbf{38.45} \\
aaditya/Llama3-OpenBioLLM-70B & 61.33 & \textbf{65.92} & 41.77 & \textbf{45.89} \\
johnsnowlabs/JSL-MedLlama-3-8B-v2.0 & 35.98 & \textbf{54.09} & 31.33 & \textbf{39.44} \\
\hline
Qwen/Qwen2.5-72B-Instruct & 76.87 & \textbf{81.93} & 58.35 & \textbf{63.33} \\
Qwen/Qwen2.5-7B-Instruct & 51.92 & \textbf{67.73} & 43.71 & \textbf{48.38} \\
meta/llama-Meta-Llama-3.1-8B-Instruct & 51.25 & \textbf{64.69} & 41.06 & \textbf{47.71} \\
meta/llama-Meta-Llama-3.1-70B-Instruct & 80.94 & \textbf{82.60} & 61.75 & \textbf{63.17} \\
meta/llama-Llama-3.2-3B-Instruct & 39.22 & \textbf{52.08} & 32.16 & \textbf{36.87} \\
mistralai/Mistral-Small-Instruct-2409 & 51.72 & \textbf{63.70} & 40.90 & \textbf{43.47} \\
mistralai/Mistral-Large-Instruct-2407 & 76.75 & \textbf{82.40} & 56.29 & \textbf{62.14} \\
\hline
speakleash/Bielik-11B-v2.2-Instruct & \textbf{62.36} & 56.98 & \textbf{43.20} & 42.88 \\
\hline
gpt-4o-mini-2024-07-18 & 75.88 & \textbf{75.92} & 54.94 & \textbf{56.88} \\
gpt-4o-2024-08-06 & \textbf{89.96} & 88.69 & \textbf{73.89} & 72.51 \\
\hline
\end{tabular}
\caption{The comparison of LLMs on Polish and English datasets, using the same LEK and LDEK exams, is represented as a percentage of correct answers.}
\label{tab:resultsenpercents}
\end{table*}

\begin{table*}[h!]
\centering
\small
\begin{tabular}{l|rr|rr}
\hline
\textbf{Model Name} & \textbf{LEK} & \textbf{LEK (en)} & \textbf{LDEK} & \textbf{LDEK (en)} \\
\hline
BioMistral-BioMistral-7B & 0/13 & 0/13 & 0/13 & 0/13 \\
OpenMeditron-Meditron3-8B & 0/13 & \textbf{13/13} & 0/13 & 0/13 \\
OpenMeditron-Meditron3-70B & \textbf{13/13} & \textbf{13/13} & 0/13 & 0/13 \\
ProbeMedicalYonseiMAILab-medllama3-v20 & 0/13 & \textbf{3/13} & 0/13 & 0/13 \\
aaditya-Llama3-OpenBioLLM-70B & \textbf{13/13} & \textbf{13/13} & 0/13 & 0/13 \\
johnsnowlabs-JSL-MedLlama-3-8B-v2.0 & 0/13 & \textbf{4/13} & 0/13 & 0/13 \\
\hline
Qwen-Qwen2.5-72B-Instruct & \textbf{13/13} & \textbf{13/13} & 11/13 & \textbf{13/13} \\
Qwen-Qwen2.5-7B-Instruct & 2/13 & \textbf{13/13} & 0/13 & 0/13 \\
meta-llama-Meta-Llama-3.1-8B-Instruct & 1/13 & \textbf{13/13} & 0/13 &  0/13 \\
meta-llama-Meta-Llama-3.1-70B-Instruct & \textbf{13/13} & \textbf{13/13} & 12/13 & \textbf{13/13} \\
meta-llama-Llama-3.2-3B-Instruct & 0/13 & \textbf{2/13} & 0/13 & 0/13 \\
mistralai-Mistral-Small-Instruct-2409 & 1/13 & \textbf{13/13} & 0/13 & 0/13 \\
mistralai-Mistral-Large-Instruct-2407 & \textbf{13/13} & \textbf{13/13} & 8/13 & \textbf{13/13} \\
\hline
speakleash-Bielik-11B-v2.2-Instruct & \textbf{13/13} & 8/13 & 0/13 & 0/13 \\
\hline
gpt-4o-mini-2024-07-18 & \textbf{13/13} & \textbf{13/13} & 6/13 & \textbf{9/13} \\
gpt-4o-2024-08-06 & \textbf{13/13} & \textbf{13/13} & \textbf{13/13} & \textbf{13/13} \\
\hline
\end{tabular}
\caption{The comparison of LLMs on Polish and English datasets using the same LEK and LDEK exams is represented as a passed exams.}
\label{tab:resultsenspassed}
\end{table*}

\section{Comparison against human results}
\label{sec:humanresults} 

\texttt{Meditron3-70B}, \texttt{Meta-Llama-3.1-70B-Instruct}, \texttt{Bielik-11B-v2.2-Instruct}, and \texttt{gpt-4o-2024-08-06} are selected as the top-performing models for the groups mentioned in Section \ref{sec:perfrormance}, and compared against anonymized human results published on the CEM webpage from the last four LEK and LDEK sessions (Spring 2024, Autumn 2023, Spring 2023, Autumn 2022), covering 977 LEK and 984 LDEK questions. The exams were taken by 33,929 participants for LEK and 4,366 for LDEK, totaling 38,295 results from medical graduates and final-year students in Poland. While all selected models pass the chosen LEK exams, only \texttt{Meta-Llama-3.1-70B-Instruct} and \texttt{gpt-4o-2024-08-06} score within the range defined by an average number of points $\pm$ standard deviation achieved by humans. Assuming a normal distribution of exam results, it could be concluded that these models perform as a typical medical student. Notably, for the spring 2024 LEK exam, \texttt{Meditron3-70B} also achieves an average-level result, while \texttt{gpt-4o-2024-08-06} exceeds the average student score. For the LDEK exams, all models perform noticeably worse. Assuming a normal distribution of exam results, only \texttt{gpt-4o-2024-08-06} maintains a performance level comparable to that of an average medical student, consistent with its LEK exam results. In contrast, \texttt{Meditron3-70B} and \texttt{Bielik-11B-v2.2-Instruct} perform poorly, failing all exams, while \texttt{Meta-Llama-3.1-70B-Instruct} score below the average but manage to pass each exam. These outcomes are summarized in Table \ref{tab:humansvsllm}.

The same models are used to compare their performance with humans on the PES exams. More details about joining the PES medical questions and human results are provided in Appendix \ref{appendix:human_vs_llm}. The best-performing model is \texttt{gpt-4o-2024-08-06}, which achieves results in above 60\% of cases better than half of the test takers population and above 30\% of cases is placed in the top 25\% of scores. Notably, this model outperforms all examinees in a thoracic surgery exam. However, it is important to note that the examinee population for this particular exam is relatively small, consisting of only six participants. However, it is worth noting that even the best model achieves results worse than half of the test takers population in over 30\% of specializations. For the \textit{Audiology \& phoniatrics} specialization, the model underperformes compared to all examinees. However, the test takers population for that particular case was relatively small, consisting of only nine participants. The second-best model, \texttt{Meta-Llama-3.1-70B-Instruct}, delivers significantly worse performance compared to the best model. Only slightly above 11\% of its results across specializations are above the population median, while in over 30\% of medical specializations, its performance is above the 25th percentile and below the 50th percentile. The remaining models, \texttt{Meditron3-70B} and \texttt{Bielik-11B-v2.2-Instruct}, perform extremely poorly, with most of their results falling below the 25th percentile or even below the lowest scores of the entire test takers population. The human results and additional explanations for Table \ref{tab:pes_human_llm_details} are provided in Appendix \ref{appendix:human_vs_llm}, where whiskers indicate the minimum and maximum human scores rather than the inter-quartile range.

\begin{table*}[htbp]
\small
\centering
\small

\begin{tabular}{l|ccc}
\hline
\textbf{Model Name} & \textbf{Criteria} & \textbf{Number of cases} & \textbf{Percentage share} \\
\hline
\multirow{6}{*}{OpenMeditron/Meditron3-70B} & $ Y < min(X) $ & 17 & 25.00\% \\
& $ Y \in [min(X), p_{25}) $ & \textbf{47} & \textbf{69.12\%} \\
& $ Y \in [p_{25}, p_{50}) $ & 2 & 2.94\% \\
& $ Y \in [p_{50}, p_{75}) $ & 2 & 2.94\%  \\
& $ Y \in [p_{75}, max(X)) $ & 0 & 0\% \\
& $ Y \ge max(X) $ & 0 & 0\% \\
\hline
\multirow{6}{*}{meta-llama/Meta-Llama-3.1-70B-Instruct} & $ Y < min(X) $ & 5 & 7.35\% \\
& $ Y \in [min(X), p_{25}) $ & \textbf{33} & \textbf{48.53\%} \\
& $ Y \in [p_{25}, p_{50}) $ & 22 & 32.35\% \\
& $ Y \in [p_{50}, p_{75}) $ & 6 & 8.83\% \\
& $ Y \in [p_{75}, max(X)) $ & 2 & 2.94\% \\
& $ Y \ge max(X) $ & 0 & 0\% \\
\hline
\multirow{6}{*}{speakleash/Bielik-11B-v2.2-Instruct} & $ Y < min(X) $ & 22 & 33.82\% \\
& $ Y \in [min(X), p_{25}) $ & \textbf{44} & \textbf{64.71\%} \\
& $ Y \in [p_{25}, p_{50}) $ & 1 & 1.47\% \\
& $ Y \in [p_{50}, p_{75}) $ & 0 & 0\% \\
& $ Y \in [p_{75}, max(X)]) $ & 0 & 0\% \\
& $ Y \ge max(X) $ & 0 & 0\% \\
\hline
\multirow{6}{*}{gpt-4o-2024-08-06} & $ Y < min(X) $ & 1 & 1.47\% \\
& $ Y \in [min(X), p_{25}) $ & 9 & 13.24\% \\
& $ Y \in [p_{25}, p_{50}) $ & 15 & 22.06\% \\
& $ Y \in [p_{50}, p_{75}) $ & 20 & 29.41\% \\
& $ Y \in [p_{75}, max(X)) $ & \textbf{22} & \textbf{32.35\%} \\
& $ Y \ge max(X) $ & 1 & 1.47\% \\
\hline
\end{tabular}
\caption{Aggregated PES exam results categorizing model $Y$ performance relative to the test takers population $X$ across various percentiles, from scores below all examinees ($Y < min(X)$) to scores compared to or exceeding the best human results ($ Y \ge max(X) $). Additional explanations are available in Appendix \ref{appendix:human_vs_llm}.}
\label{tab:pes_human_llm_details}
\end{table*}

\begin{table*}[htbp]
    \centering
    \small
    \subfigure[LEK]{
\begin{tabular}{l|ccccc}
\hline
\textbf{Model / Human} & \textbf{2024S} & \textbf{2023A} & \textbf{2023S} & \textbf{2022A}\\
\hline
OpenMeditron/Meditron3-70B & 153 & \color{orange} 133 & \color{orange} 130 & \color{orange}125 \\
\hline
meta-llama/Meta-Llama-3.1-70B-Instruct & 170 & 162 & 153 & 161 \\
\hline
speakleash/Bielik-11B-v2.2-Instruct & \color{orange} 129 & \color{orange} 122 & \color{orange} 123 & \color{orange} 133 \\
\hline
gpt-4o-2024-08-06 & \color{teal}184 & 177 & 176 & 179 \\
\hline
Average human result & 163.47 & 163.36 & 161.11 & 165.64 \\
with standard deviation & $\pm19.79$ & $\pm18.38$ & $\pm18.66$ & $\pm16.95$\\
\hline
\end{tabular}
\label{tab:lekhumanvsllms}
}
    \\ 
    \subfigure[LDEK]{
\begin{tabular}{l|ccccc}
\hline
\textbf{Model / Human} & \textbf{2024S} & \textbf{2023A} & \textbf{2023S} & \textbf{2022A}\\
\hline
OpenMeditron/Meditron3-70B & \color{red} 103 & \color{red} 83 & \color{red} 94 & \color{red} 95 \\
\hline
meta-llama/Meta-Llama-3.1-70B-Instruct & \color{orange} 121 & \color{orange} 119 & \color{orange} 124 & \color{orange} 123 \\
\hline
speakleash/Bielik-11B-v2.2-Instruct & \color{red} 100 & \color{red} 74 & \color{red} 83 & \color{red} 85 \\
\hline
gpt-4o-2024-08-06 & 139 & 136 & 144 & 136 \\
\hline
Average human result & 147.62 & 148.57 & 149.42 & 156.22 \\
with standard deviation & $\pm26.08$ & $\pm19.08$ & $\pm21.13$ & $\pm23.52$\\
\hline
\end{tabular}
\label{tab:ldekhumanvsllms}
    }
    \caption{Comparison of top-performing LLMs and average human results, including standard deviation, across selected LEK and LDEK exams. Red represents values below the passing threshold of 112 points, orange highlights scores below average minus one standard deviation, green indicates scores above average plus one standard deviation, and black represents scores within one standard deviation of the average.}
    \label{tab:humansvsllm}
\end{table*}

\section{Conclusion}
In this paper, we propose a new benchmark for analyzing the performance of large language models in answering questions pertaining to the domain of medical knowledge. 
In contrast to the majority of previous medical datasets that collect examination questions in English, our dataset is derived from data of Polish origin.
We show that general-purpose LLMs, trained on internet-scale datasets with extensive computational resources, outperform medical-specific models and that using a general-purpose model fine-tuned specifically for the Polish language is justified only if models of a similar size are considered.
%


LLMs performance varies across different medical exams and languages. Most models are able to pass the LEK exam, but many struggle with the LDEK exam. In the case of the PES exams for various medical specializations, the performance is even lower, with only \texttt{gpt-4o-2024-08-06} maintaining satisfactory results. However, even the top-performing model scores lower than at least half of the test takers in over 30\% of specializations. This highlights the need for thorough verification before implementing LLMs in the medical domain, as the results are not consistently reliable across all medical specialties.

The parallel sub-corpus composed of examination questions in Polish aligned with their English counterparts is a distinguished feature of the presented benchmark which allows us to investigate the cross-lingual transfer of medical knowledge in LLMs. Our findings show that models perform significantly better on English questions and that as the size of the model increases, performance improves, and the gap between languages narrows, an expected but difficult-to-measure result without an appropriate benchmark.

\clearpage

\section*{Limitations}


While LLMs have demonstrated impressive performance on Polish medical multiple-choice exams, this achievement represents only a narrow facet of medical expertise. Becoming a licensed physician in Poland requires extensive training, rigorous coursework, and hands-on experience with practical medical procedures—far beyond what written exams can assess. Clinical practice necessitates analyzing diverse information and solving complex problems with multiple possible solutions. Physicians must determine what data is needed, obtain it through patient interviews, physical examinations, diagnostic tests, and consultations—all heavily reliant on direct human interaction that AI models cannot replicate. Moreover, the exams are multiple-choice, and real-world work is not narrowed to a few possible options. Therefore, despite strong exam results, LLMs cannot currently substitute the comprehensive qualifications and essential human interactions integral to effective medical care. However, this study demonstrates that LLMs may serve as valuable tools for medical practitioners, a potential use case previously suggested by other researchers \cite{ullah2024challenges,park2024assessing,clark2024chatbots,liu123,doi:10.1056/NEJMsr2214184}.

Due to regional access restrictions, we were unable to evaluate PaLM 2 \cite{palm2} and certain Llama 3.2 models. Additionally, highly resource-intensive models such as \texttt{Meta-Llama-3.1-405B-Instruct} or some other restricted access LLMs, such as Gemini \cite{gemini} were not evaluated.

The presented research and provided benchmark are primarily designed to evaluate the performance of LLM models in medical question-answering tasks in the Polish language, serving as a substitute for the MedQA benchmark. To prevent models from being trained on the benchmark data, the training dataset is not provided. The LEK and LDEK exams have a similar number of examples in the final dataset, but the PES exam questions make up the majority. While the English questions for the LEK and LDEK exams are in the minority, there are no PES questions in English. This benchmark is not intended for other NLP tasks such as text classification, named entity recognition, sentiment analysis, or other related problems.

The GPT-4o model accessed via the OpenAI API was used for our analysis. The web search option was not enabled, ensuring that the model did not actively search the Internet for answers to the medical questions asked.

The dataset described in this paper was collected from an examination center's webpage, where the questions are freely available. These exams can be used, for example, by medical students preparing for their assessments. There is a potential risk for these exams being included in the training datasets of evaluated LLMs. 
Therefore, the evaluation results presented in this paper must be treated with the same degree of caution as the results determined with the use of any other publicly available dataset such as MMLU \cite{mmlu} or scores reported on leaderboards that aggregate results determined for publicly available datasets, such as Open LLM Leaderboard \cite{open-llm-leaderboard-v2}.
However, taking into consideration that our dataset originates from a highly authoritative source, creating a dataset of comparable size and quality from the ground up would be prohibitively difficult.





\bibliography{anthology,custom}
\bibliographystyle{acl_natbib}

\appendix

\section{Polish medical exams detailed description}
\label{appendix:detailedexams}
Medical studies in Poland last 6 years, while dentistry takes 5 years. Final-year students and graduates can take their respective final exams — LEK for medicine and LDEK for dentistry. Passing the final examination and completing a postgraduate internship are required to obtain a medical license.\footnote{\url{https://www.cem.edu.pl/lek_info.php}\newline\url{https://www.cem.edu.pl/ldek_info.php}  }
\hypersetup{breaklinks=true}
Both LEK and LDEK are four-hour exams conducted twice a year. Each exam consists of 200 multiple-choice questions with five possible answers, of which only one is correct. The questions cover a wide range of medical or dental disciplines. The distribution of questions from various fields is presented in Tables \ref{tab:t1}  and \ref{tab:t2}. To pass, a candidate must correctly answer at least 56\% of the questions. Physicians and dentists can retake these exams multiple times, even after passing, if they are dissatisfied with their score.\footnote{\url{https://www.cem.edu.pl/lek_info.php}\newline
\url{https://www.cem.edu.pl/ldek_info.php}} A controversial rule (\url{https://pulsmedycyny.pl/kadry/lekarze/samorzad-lekarski-postuluje-pilna-} \url{zmiane-bazy-pytan-w-lek-i-ldek/}) has been introduced in 2022, stipulating that 70\% of the exam questions come from a publicly available database, which includes 2,870 questions for LEK and 3,198 for LDEK. After these changes, the average exam scores and the percentage of passing candidates increased significantly.\footnote{\url{https://www.cem.edu.pl/lep_s_h.php}\newline \url{https://www.cem.edu.pl/ldep_s_h.php}}

\begin{table}[h]
\centering
\small
\begin{tabular}{lr}
\hline
\textbf{Discipline} & \textbf{Questions}\\
\hline
Internal medicine* & 39 \\
Pediatry* & 29 \\
Surgery* & 27 \\
Obstetrics and gynecology* & 26 \\
Psychiatry & 14 \\
Family medicine* & 20 \\
Emergency medicine and intensive care & 20 \\
Bioethics and medical law & 10 \\
Medical certification & 7 \\
Public health & 8 \\
\hline
\end{tabular}
\caption{Distribution of test questions in LEK. The disciplines marked with an asterisk contribute to a minimum of 30 oncology-related questions. Internal medicine includes cardiovascular diseases. Pediatry includes neonatology. Surgery includes trauma surgery.}
\label{tab:t1}
\end{table}

\begin{table}[h]
\centering
\small
\begin{tabular}{lr}
\hline
\textbf{Discipline} & \textbf{Questions}\\
\hline
Conservative dentistry* & 46 \\
Pediatric dentistry* & 29 \\
Oral surgery* & 25 \\
Prosthetic dentistry & 25 \\
Periodontology* & 20 \\
Orthodontics* & 20 \\
Emergency medicine & 10 \\
Bioethics and medical law & 10 \\
Medical certification & 7 \\
Public health & 8 \\
\hline
\end{tabular}
\caption{Distribution of test questions in LDEK. The disciplines marked with an asterisk contribute to a minimum of 25 oncology-related questions.}
\label{tab:t2}
\end{table}

The PES exam is available to physicians and dentists who have completed the required internships and courses as part of their specialization training. Passing PES is mandatory to obtain the title of a specialist in a medical field. The exam consists of two parts: a written test and an oral examination. It is typically held twice a year for each medical specialty. The duration of the written test varies depending on the specialty, but it generally consists of 120 multiple-choice questions with five possible answers, of which one is correct. A minimum of 60\% correct answers are required to pass. Unlike LEK and LDEK, none of the PES questions are public before the exam. Candidates who score at least 70\% on the written test are exempt from taking the oral part of the exam, a rule implemented at the end of 2022. The format of the oral (practical) exam varies by specialty\footnote{\url{https://www.cem.edu.pl/spec.php}}. PES is generally considered to be the most challenging knowledge verification in the whole career of a medical doctor in Poland.

\section{Example exam questions}
\label{appendix:examplequestions}
\subsection{LEK}
Exam: 2022 Spring \\
Question id: 77 
\begin{verbatim}
Przepuklina u starszego mężczyzny
z chorobą obturacyjną płuc uwypuklająca 
się na zewnątrz jamy brzusznej przez 
powięź poprzeczną stanowiącą tylną 
ścianę kanału pachwinowego w miejscu 
ograniczonym od góry przez ścięgno 
łączące, od dołu przez więzadło 
pachwinowe, a bocznie przez naczynia 
nabrzuszne dolne - jest rozpoznawana 
jako:
A. przepuklina pachwinowa skośna.
B. przepuklina mosznowa.
C. przepuklina pachwinowa prosta.
D. przepuklina udowa.
E. przepuklina Spigela.    
\end{verbatim}
Correct answer: C.
\subsection{LEK (en)}
This sample is a translation of the above question (LEK) provided by the examination center. \\ \ \\
Exam: 2022 Spring \\
Question id: 77 
\begin{verbatim}
An elderly male patient with obturative 
lung disease was diagnosed with hernia. 
It was protruding from the abdominal 
cavity through the transverse fascia 
which forms the posterior wall of the 
inguinal canal, at the site bordering 
the conjoint tendon at the top, the 
inguinal ligament at the bottom, and 
laterally, through inferior  epigastric 
vessels. The hernia in such location is 
known as:
A. oblique inguinal hernia.
B. scrotal hernia.
C. direct inguinal hernia.
D. femoral hernia.
E. spigelian hernia.
\end{verbatim}
Correct answer: C.
\subsection{LDEK}
Exam: 2022 Spring \\
Question id: 77 
\begin{verbatim}
Jednostronny wyciek z nosa posokowatej 
treści z domieszką krwi, rozchwianie 
zębów górnych, łzawienie, wytrzeszcz 
gałki ocznej, a niekiedy bóle i 
mrowienie policzka mogą być wczesnym 
objawem:
A. pseudotorbieli zatoki szczękowej.
B. raka zatoki szczękowej.
C. raka policzka.
D. przewlekłego zapalenia zatoki szczękowej.
E. ostrego zapalenia zatoki szczękowej.
\end{verbatim}
Correct answer: B.
\subsection{LDEK (en)}
This sample is a translation of the above question (LDEK) provided by the examination center. \\ \ \\
Exam: 2022 Spring \\
Question id: 77 
\begin{verbatim}
Unilateral ichorous discharge from the 
nose with a blend of blood, gomphiasis 
of the upper teeth, lacrimation, 
exopathalmos, and sometimes pain and 
tingling sensation in the cheek, might 
be an early symptom of:
A. pseudocyst of the maxillary sinus.
B. cancer of the maxillary sinus.
C. buccal cancer.
D. chronic maxillary sinusitis.
E. acute maxillary sinusitis.
\end{verbatim}
Correct answer: B.
\subsection{PES}
Exam: 2019 Autumn \\
Question id: 68 \\ 
Specialty: Family medicine
\begin{verbatim}
 Kliniczne cechy sepsy u dzieci to:
1) gorączka;
2) leukocytoza;
3) leukopenia;
4) tachykardia bez innej przyczyny;
5) tachypnoe bez innej przyczyny.
\end{verbatim}
Correct answer: E.

\section{Prompts}
\label{appendix:prompts}

\subsection{Prompt in Polish}
\begin{verbatim}
Twoje zadanie to udzielenie odpowiedzi
na test medyczny dla lekarzy. Spośród
wszystkich odpowiedzi A,B,C,D,E wybierz
tylko jedną. Jeżeli nie jesteś pewien,
wybierz najbardziej prawdopodobną.
Odpowiedz w sposób:
Prawidłowa odpowiedź to B.
\end{verbatim}

\subsection{Prompt in English}
\begin{verbatim}
Your task is to answer a medical test
for doctors. From all the options
A, B, C, D, E, choose only one. If you're
unsure, select the most probable one.
Respond in the following manner:
The correct answer is B.
\end{verbatim}

\section{Specialty performance on PES}
\label{appendix:specialities}

Among the 72 unique PES specialties, certain areas of medicine consistently challenge the majority of tested models, while others frequently rank among the highest-scoring categories based on model accuracy. By identifying the top five highest and lowest-scored categories, we gain insights into specific domains where models excel or struggle, highlighting their potential limitations in these fields. 

The general field of medicine where LLMs struggle the most is dentistry, specifically in orthodontics, which appeared ten times in the top five lowest scores across 17 models, followed by conservative dentistry with endodontics and pediatric dentistry. These results suggest that certain nuances in dental specialties may not yet be fully captured by modern LLMs, leading to difficulties in understanding this broad field.

The most frequently occurring specialty among the highest-scoring categories was laboratory diagnostics, which appeared twelve times. This observation may indicate that diagnostics tasks align well with the pattern recognition and data interpretation capabilities of LLMs.
Additionally, other specialties with high scores, such as public health and pulmonary diseases reflect the vast quantity and accessibility of data in those fields. The COVID-19 pandemic could have largely increased the resource pool regarding pulmonary and respiratory conditions.

\begin{figure}[h]
    \centering
    \includegraphics[width=.5\textwidth]
    {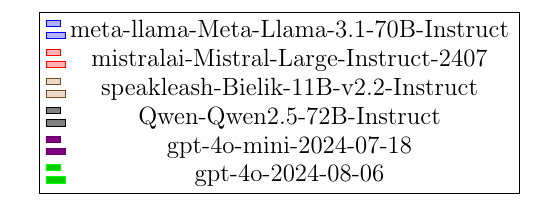}
    \label{fig:graph_desc}
\end{figure}

\clearpage
\begin{figure*}[h]
    \centering
    \includegraphics[width=1.0\textwidth]{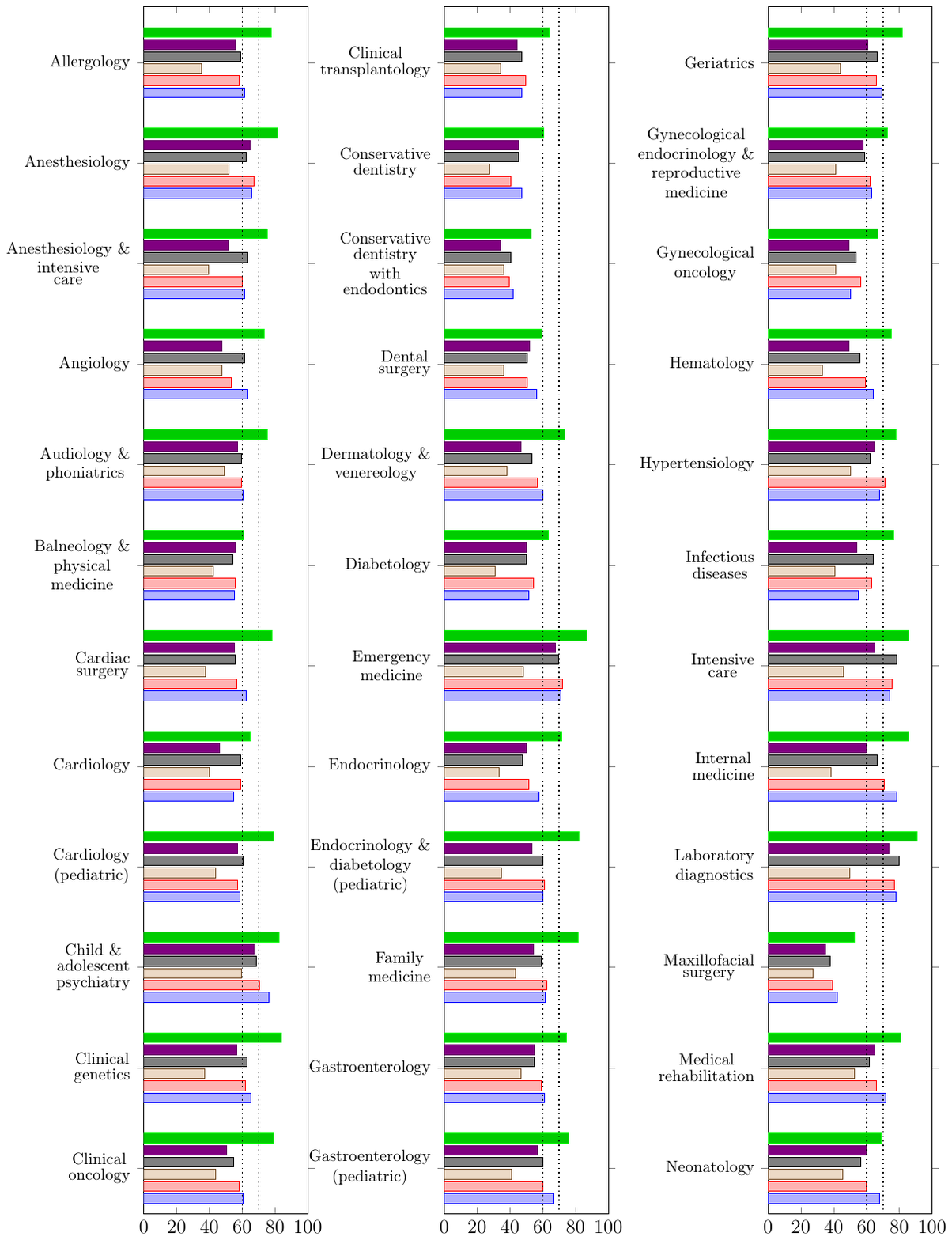}
    \caption{Models performance on different specialties on PES exams (part 1/2). Dotted lines indicate the passing threshold for the exam (60\%) and exemption from the oral part (75\%).}
    \label{fig:graph1}
\end{figure*}

\clearpage
\begin{figure*}[h]
    \centering
    \includegraphics[width=1.0\textwidth]
    {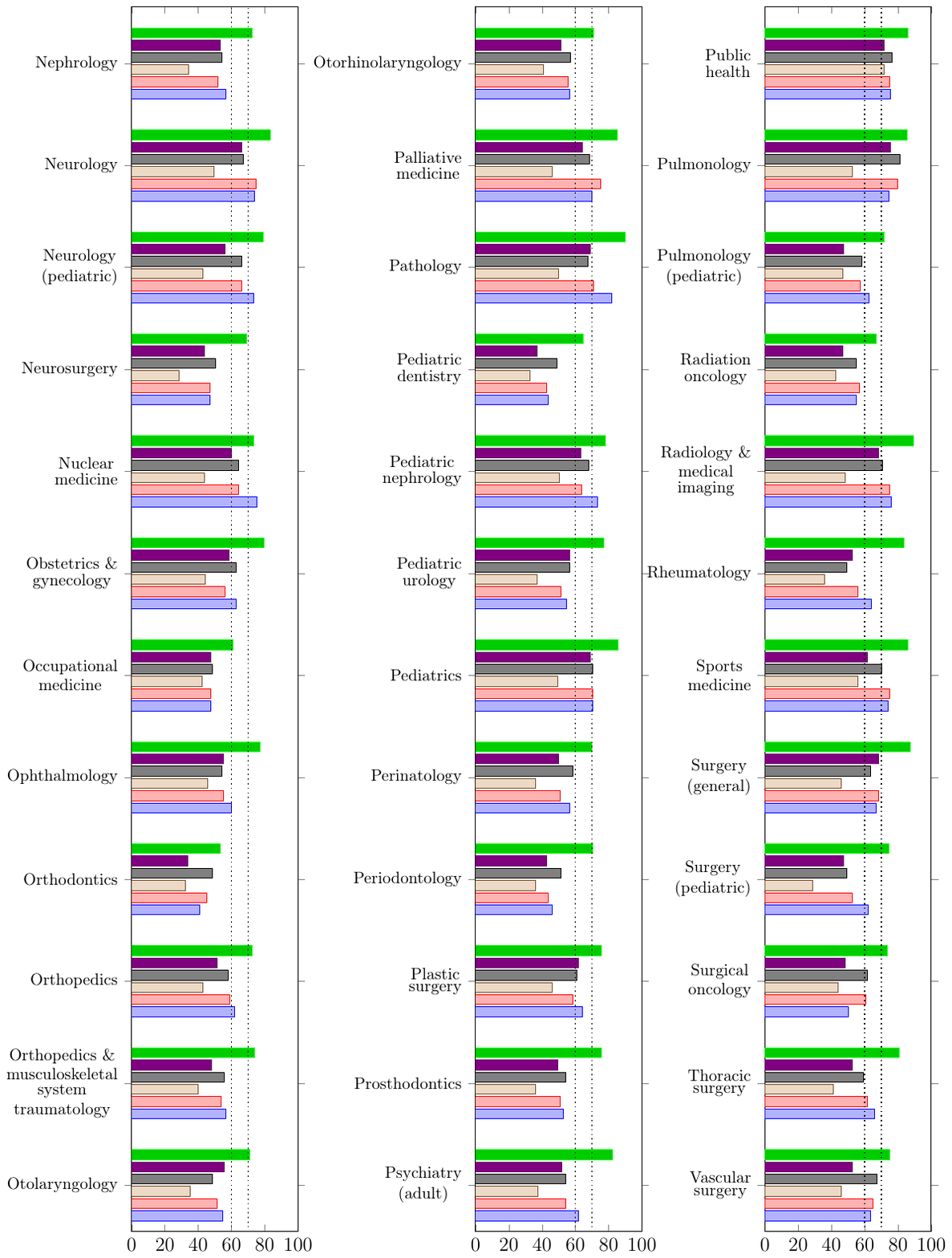}
    \caption{Models performance on different specialties on PES exams (part 2/2). Dotted lines indicate the passing threshold for the exam (60\%) and exemption from the oral part (75\%).}
\end{figure*}
\clearpage

\section{Data preparation} \label{appendix:data_prep}
\subsection{Data sources}
Medical exams in Poland are conducted biannually, in spring and autumn. Past exam content and corresponding answers are available on the \href{https://cem.edu.pl/index.php}{Medical Examination Center} (Centrum Egzaminów Medycznych, CEM) website, either as quizzes or PDF files. The site archives the following exams in the Polish language:
\begin{itemize}
    \item LEK exams from autumn 2008 to autumn 2012 are provided as PDF files,
    \item LEK exams from spring 2013 to autumn 2015, and from spring 2021 to autumn 2024 are available as quizzes,
    \item LDEK exams from autumn 2008 to autumn 2012 are available as PDF files,
    \item LDEK exams from spring 2013 to autumn 2015, and from spring 2021 to autumn 2024 are provided as quizzes,
    \item PES exams from spring 2003 to autumn 2017, and from spring 2023 to spring 2024 are available as quizzes.
\end{itemize}
LEK and LDEK exams published as quizzes are also available in English. The missing LEK and LDEK exams from spring 2016 to autumn 2020 have not been found. The missing PES exams from spring 2018 to autumn 2022 have been published as PDF files on the \href{https://nil.org.pl/}{Supreme Medical Chamber} (Naczelna Izba Lekarska, NIL) website. 

\begin{figure}[h]
    \centering
    \includegraphics[width=1\linewidth]{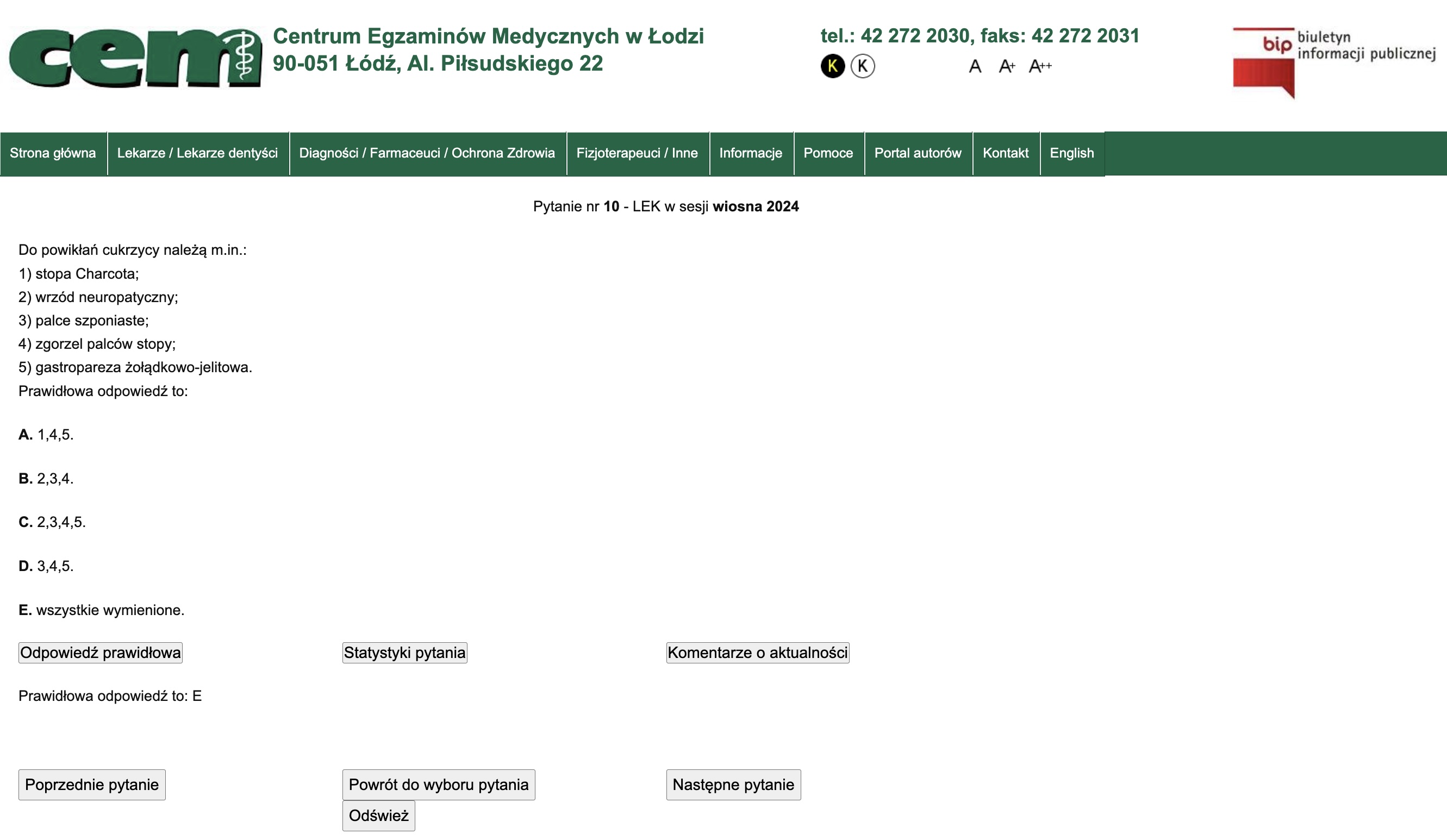}
    \caption{Quiz interface on the Medical Examination Center website.}
    \label{fig:quize-interface}
\end{figure}

The Medical Examination Center also provides detailed information about human answers for the PES exams. The initial view displays a list of examinees, represented by code numbers, along with their total achieved points and final grades. For all exams conducted since autumn 2006, detailed answers for each examinee are available by clicking on the examinee's code number. This detailed view includes the question number, the answer provided, and the correct answer. For the LEK and LDEK exams, only aggregated statistics of human results are published on the Medical Examination Center’s website. These include overall summary numbers, statistics broken down by university, and data grouped by specific categories, such as individuals who completed their studies within the last two years, those who graduated more than two years ago, first-time test-takers, and more. Unfortunately, these groupings are not consistent over the years. Therefore, only general aggregated statistics - such as minimum, maximum, average, standard deviation, the number of passes, the number of fails, the number of exam takers, and the number of registered candidates—can be considered reliably useful.

\subsection{Data acquisition and processing}
The missing PES exams were published on the Supreme Medical Chamber platform across two distinct pages, with separate archives for the periods 2018–2020 and 2021–2022. Each medical specialization’s exams were compressed into a zip file and provided as individual download links. To streamline the downloading process, a JavaScript script was executed via Chrome's Developer Tools, iterating through the links and simulating clicks for automatic downloads. The exams were then categorized by specialization, with each folder containing two types of PDF files: questions and the corresponding correct answers.

Custom Python scraping scripts were developed to automate the downloading of quizzes from the Medical Examination Center platform. Separate scripts were created for LEK/LDEK exams, PES exams, and exam statistics. Due to the server's slow response time, the entire process took several days, even with parallelized data download. When too many concurrent threads were used, the server became overwhelmed, resulting in timeouts.

\begin{figure}[h]
    \centering
    \includegraphics[width=1\linewidth]{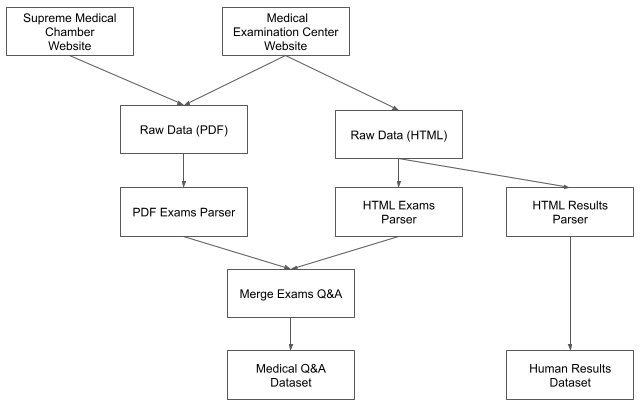}
    \caption{Data acquisition and processing workflow}
    \label{fig:data-processing}
\end{figure}

\subsection{Data quality}
Data is stored in two formats: PDF and HTML, both of which are inconsistent and present several challenges. Since the goal of creating this dataset is to establish a Polish medical benchmark for Large Language Models, questions containing images were excluded. Additionally, some questions were disqualified by their authors due to errors or inconsistencies with current medical knowledge.

\subsubsection{HTML format}
HTML format is relatively straightforward to process, as specific HTML tags can be used to extract information such as questions and correct answers. However, some questions contain images that are essential for context, which poses a challenge for AI models designed to process text. Since the final dataset is intended for text-based AI models, questions containing images were excluded using specific tags. Additionally, the quiz interface allows anonymous users to leave comments on individual questions. These comments could potentially highlight areas where the content's alignment with current knowledge has been questioned. However, many of the comments appeared unprofessional and seemed not to be moderated by the platform administrators. As a result, the presence of comments was not considered a valid indicator for filtering questions, and all of them were kept in the final dataset. 

Moreover, the raw dataset contains empty questions. The platform uses two static drop-down lists to browse questions based on exam date and medical specialization, even when no corresponding exam or question is available in the database. According to the platform's messages, missing data occurs either due to the absence of questions in the database or because exams were not conducted during a specific time. This design leads to a collection of HTML files with no meaningful content. Since the user interface does not manage these cases, it was necessary to filter out and remove such files from the dataset after downloading.

\begin{figure}[h]
    \centering
    \includegraphics[width=1\linewidth]{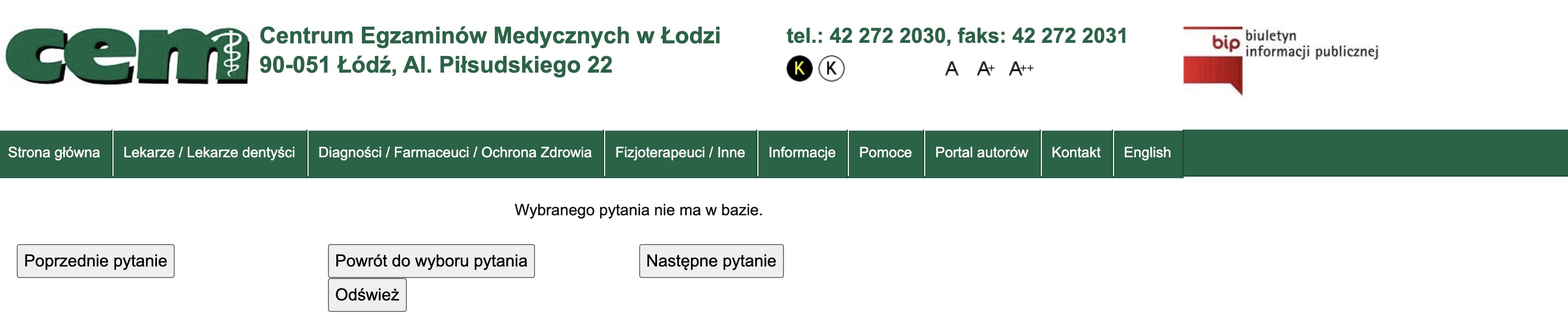}
    \caption{Example of missing data caused by an absent question.}
    \label{fig:interface_missing_question}
\end{figure}

\begin{figure}[h]
    \centering
    \includegraphics[width=1\linewidth]{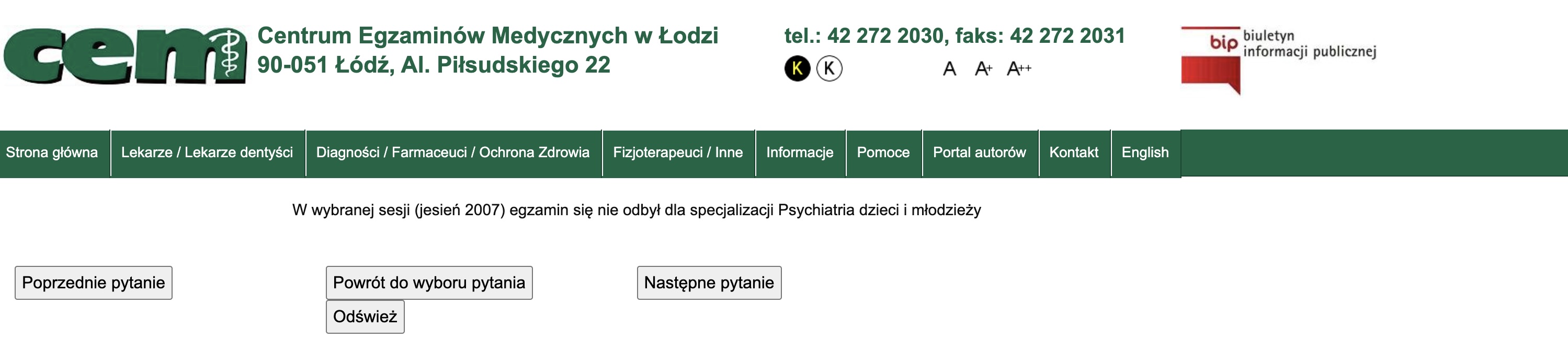}
    \caption{Example of missing data due to an exam not being conducted.}
    \label{fig:interface_exam_not_conducted}
\end{figure}

\subsubsection{PDF files}
Processing PDF files is more challenging compared to HTML due to the need to handle content sequentially, line by line, while applying multiple conditions to accurately extract medical exam questions. Additionally, the structure of questions is inconsistent across points, pages, and files. The question content or answer options may be presented in various formats, such as horizontal lists, vertical lists, two separate lists of options, or a table where points must be matched across columns. This inconsistency complicates the extraction process and poses difficulties for data processing.

\begin{figure}[h]
\centering
\hfill
\subfigure {\includegraphics[width=0.49\linewidth]{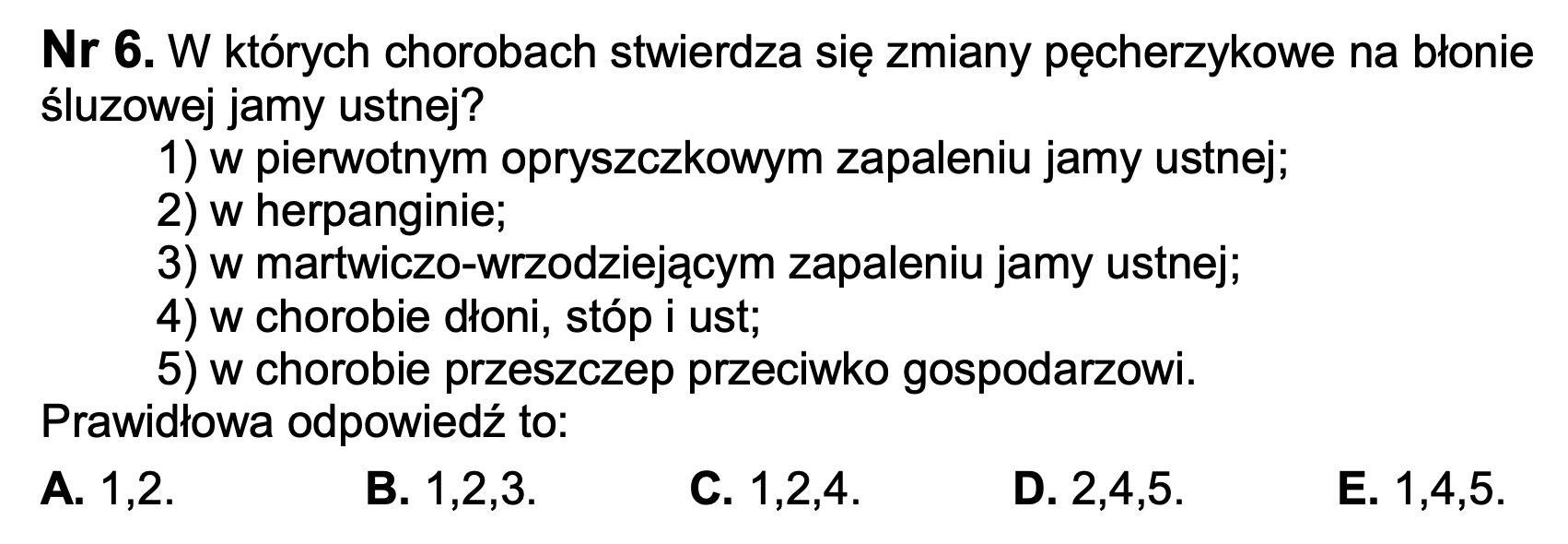}}
\hfill
\subfigure {\includegraphics[width=0.49\linewidth]{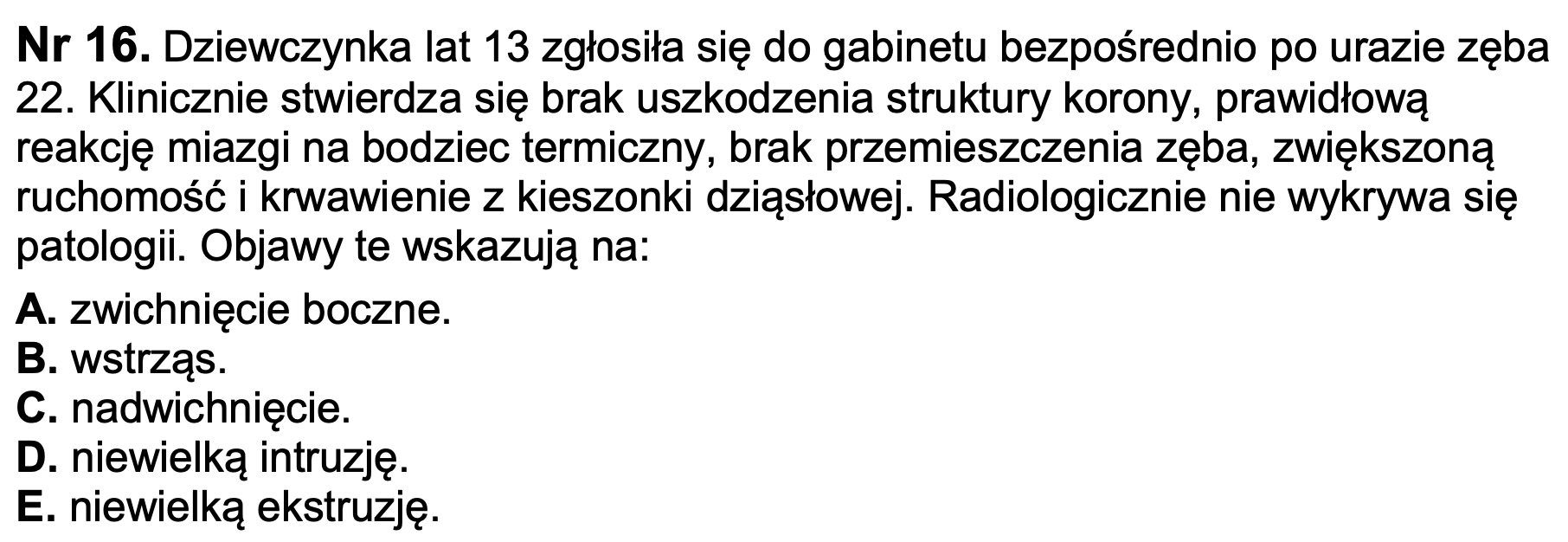}}
\hfill
\subfigure {\includegraphics[width=0.7\linewidth]{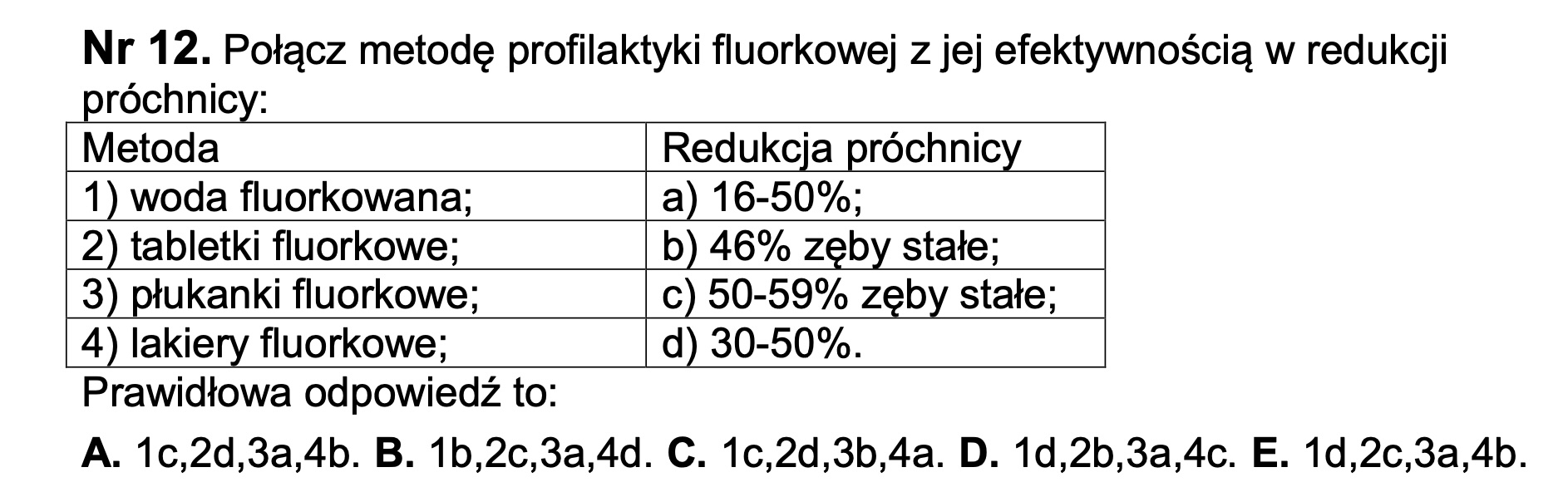}}
\hfill
\caption{Answer options presented horizontally, vertically, or in a table within the same PDF file.}
\end{figure}

\begin{table*}[ht!]
    \centering
    \resizebox{\textwidth}{!}{
    \begin{tabular}{|l|r|r|r|r|r|}
        \hline
        Model & Correct PL and EN & Incorrect (same) & Incorrect (diff) & Correct PL, Incorrect EN & Incorrect PL, Incorrect EN\\
        \hline
        OpenMeditron/Meditron3-70B & 57.93 & 18.82 & 3.99 & 10.44 & 8.82 \\
        meta-llama/Meta-Llama-3.1-70B-Instruct & 75.48 & 9.13 & 2.81 & 5.46 & 7.12 \\
        speakleash/Bielik-11B-v2.2-Instruct & 49.11 & 21.47 & 8.30 & 13.25 & 7.87 \\
        gpt-4o-2024-08-06 & 85.88 & 6.33 & 0.91 & 4.07 & 2.81 \\
        \hline
    \end{tabular}

    }
    \caption{Comparison of model results considering Polish and English responses to the same questions from \textbf{LEK} exams. For example, column \textit{Correct PL, Incorrect EN} indicates the percentage of questions answered correctly in Polish but incorrectly when the same question was translated into English, column \textit{Incorrect (diff)} indicates the percentage of questions answered incorrectly both in English and Polish, but the incorrect answers differs.}
        \label{tab:detailedcrosslinguallek}
\end{table*}

\begin{table*}[ht!]
    \centering
    \resizebox{\textwidth}{!}{
    \begin{tabular}{|l|r|r|r|r|r|}
        \hline
        Model & Correct PL and EN & Incorrect (same) & Incorrect (diff) & Correct PL, Incorrect EN & Correct EN, Incorrect PL \\
        \hline
        OpenMeditron/Meditron3-70B & 38.45 & 36.95 & 8.11 & 8.98 & 7.52 \\
        meta-llama/Meta-Llama-3.1-70B-Instruct & 52.97 & 20.13 & 7.91 & 8.78 & 10.21 \\
        speakleash/Bielik-11B-v2.2-Instruct & 33.23 & 32.52 & 14.64 & 9.97 & 9.65 \\
        gpt-4o-2024-08-06 & 66.06 & 15.31 & 4.35 & 7.83 & 6.45 \\
        \hline
    \end{tabular}
    }
        \caption{Comparison of model results considering Polish and English responses to the same questions from \textbf{LDEK} exams. For example, column \textit{Correct PL, Incorrect EN} indicates the percentage of questions answered correctly in Polish but incorrectly when the same question was translated into English, column \textit{Incorrect (diff)} indicates the percentage of questions answered incorrectly both in English and Polish, but the incorrect answers differs.}
    \label{tab:detailedcrosslingualldek}
\end{table*}

The quality of the PDF files varies significantly. While some are digitally generated with perfect clarity, others resemble scanned printed documents of noticeably lower quality. Fortunately, this variation does not impact the data extraction process. However, certain PDF files lack text layers, making them significantly harder to process, as Optical Character Recognition (OCR) must be applied to extract the text. This challenge arose for 212 exams from 2021 and 2022 year. Due to the complexity, even with OCR, it was decided to omit these documents from the analysis.

Correct answers are stored in separate PDF files. To obtain comprehensive results, content must be extracted from both the question and answer files, and the corresponding points matched. Typically, the correct answer is indicated by a letter between A and E. However, in some cases, an 'X' appears in the answer file, indicating that the question is no longer aligned with current knowledge and has been annulled.

\section{Question-level cross-lingual analysis}
\label{appendix:detailed-cross-lingual}

We select the same PL-EN dataset as in Section \ref{sec:crosslingual}. We assign each model response to a question into one of the following categories:
\begin{itemize}
    \item Correct PL and EN - a model answered correctly both to the Polish version of the question and the English translation
    \item  Incorrect (same) - a model gives incorrect answers to a question, and the answers are the same, e.g., both D (which is incorrect) for the Polish and English version
    \item Incorrect (diff) - a model gives incorrect answers to a question, and the answers are different, e.g., D for the Polish version and E for the English version
    \item Correct PL, Incorrect EN - a model gives a correct answer to a question in Polish but incorrect for the English translation
    \item Incorrect PL, Incorrect EN- a model gives an incorrect answer to a question in Polish but correct for the English translation
\end{itemize}

The results on model selection from Section \ref{sec:humanresults} are presented in Table \ref{tab:detailedcrosslinguallek} for LEK and Table \ref{tab:detailedcrosslingualldek} for LDEK.

When comparing \textit{Incorrect (same)} and \textit{Incorrect (diff)} categories, we conclude that if a model returns incorrect answers in both languages, it is more likely to produce the same incorrect answer rather than different incorrect answers for each language version. This provides strong evidence for cross-lingual knowledge transfer. However, it is also quite common for a model to answer correctly in one language while providing an incorrect response in the other.

Moreover, there are no significant differences based on language, as the proportions of 
\textit{Correct PL, Incorrect EN} and \textit{Incorrect PL, Incorrect EN} results are comparable for a given model. Even \texttt{speakleash/Bielik-11B-v2.2-Instruct} tends to generate more correct answers in Polish than in English only for the LEK dataset, though its performance remains similar across both languages in the LDEK dataset.

When analyzing all categories in which at least one language version of a question is answered incorrectly, we observe no strong preference for a specific language version. This suggests that when a model is uncertain about its response, its output may be fairly random. Based on this, we hypothesize that evaluating model outputs across different language versions could serve as a filtering mechanism to identify cases where the model has low confidence in its responses.

\section{Comparison of human results and best-performing LLMs} 
\label{appendix:human_vs_llm}

This analysis is based on a dataset derived from the intersection of human and LLM results, covering 8,062 medical questions across 68 specializations. LLM results are calculated based on the most recent exam for each specialization to ensure evaluation against up-to-date medical knowledge and minimize the impact of outdated questions. In contrast, human results are aggregated over multiple sessions to increase the sample size and improve generalizability. All human results and selected the most recent specialization questions come from 12 PES exam sessions: Spring 2024, 2023, 2018, 2017, 2016, 2012, and Autumn 2023, 2020, 2019, 2016, 2015, and 2008. Human results include 29,450 anonymized physicians and dentists in Poland who completed specialization training and took the mentioned exams.

The number of specializations and questions is smaller than in the previous analysis due to inconsistencies in specialization names across different exam years and published human results. In both human results and exam questions, specialization names sometimes vary, requiring normalization and, in some cases, the exclusion of edge cases to align both datasets.  

Table \ref{tab:pes_human_llm_details} contains an aggregated comparison between human and LLMs results for the PES exams. $X$ represents the distribution of human results, while the score of each model, $Y$, is categorized into the following ranges:
\begin{itemize}[noitemsep]
  \item $ Y < min(X) $: Indicates model $Y$ underperforms all test takers.
  \item  $ Y \in [min(X), p_{25}) $: Model $Y$ scores in the lowest 25\% of test takers.
  \item $ Y \in [p_{25}, p_{50}) $: Model $Y$ scores between the 25th and 50th percentiles, below the median but above the first quartile.
  \item $ Y \in [p_{50}, p_{75}) $: Model $Y$ scores between the median and the top 25\%.
  \item $ Y \in [p_{75}, max(X)]) $: Model $Y$ scores in the top 25\% of test takers.
  \item $ Y \ge max(X) $: Model $Y$ matches or surpasses the top human score.
\end{itemize}

\onecolumn
\begin{figure}[!h]
    \centering
    \includegraphics[width=0.9\textwidth]{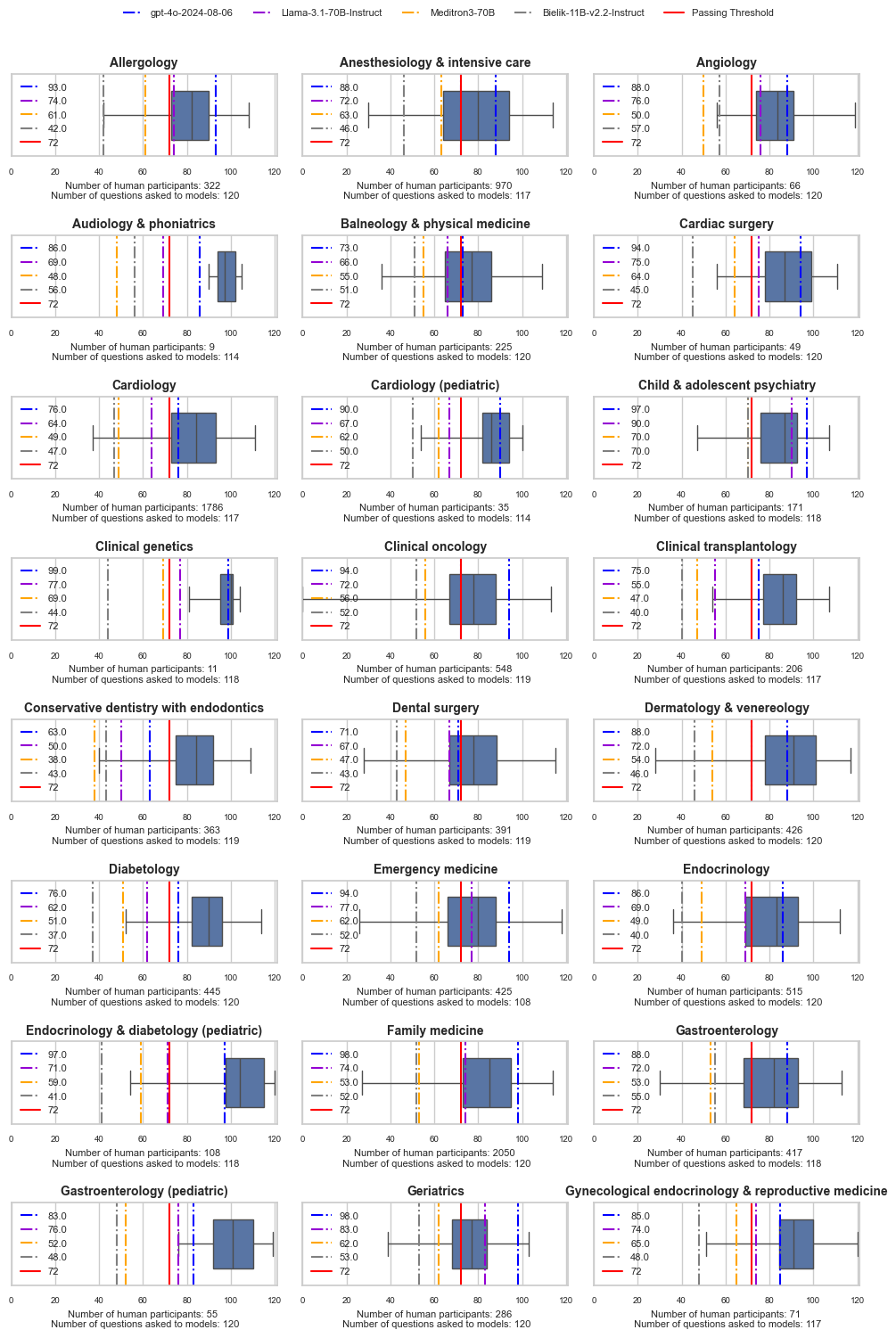}
    \caption{Students performance compared to top-performing LLMs on different specialties on PES exam (part 1/3).}    
    \label{fig:pes_human_llm_1}
\end{figure}

\begin{figure}[h]
    \centering
    \includegraphics[width=1\textwidth]{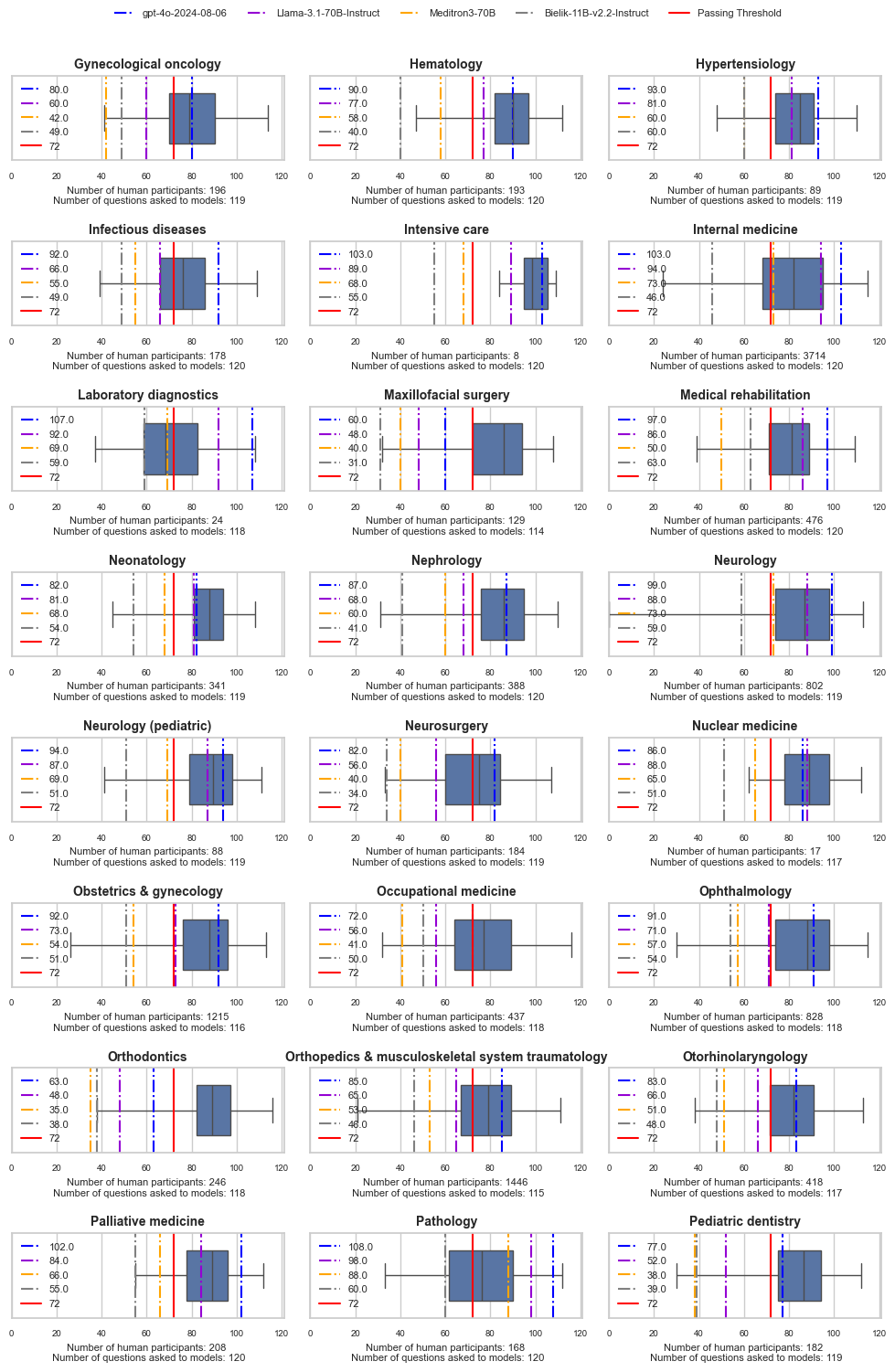}
    \caption{Students performance compared to top-performing LLMs on different specialties on PES exam (part 2/3).}
    \label{fig:pes_human_llm_2}
\end{figure}

\begin{figure}[h]
    \centering
    \includegraphics[width=1\textwidth]{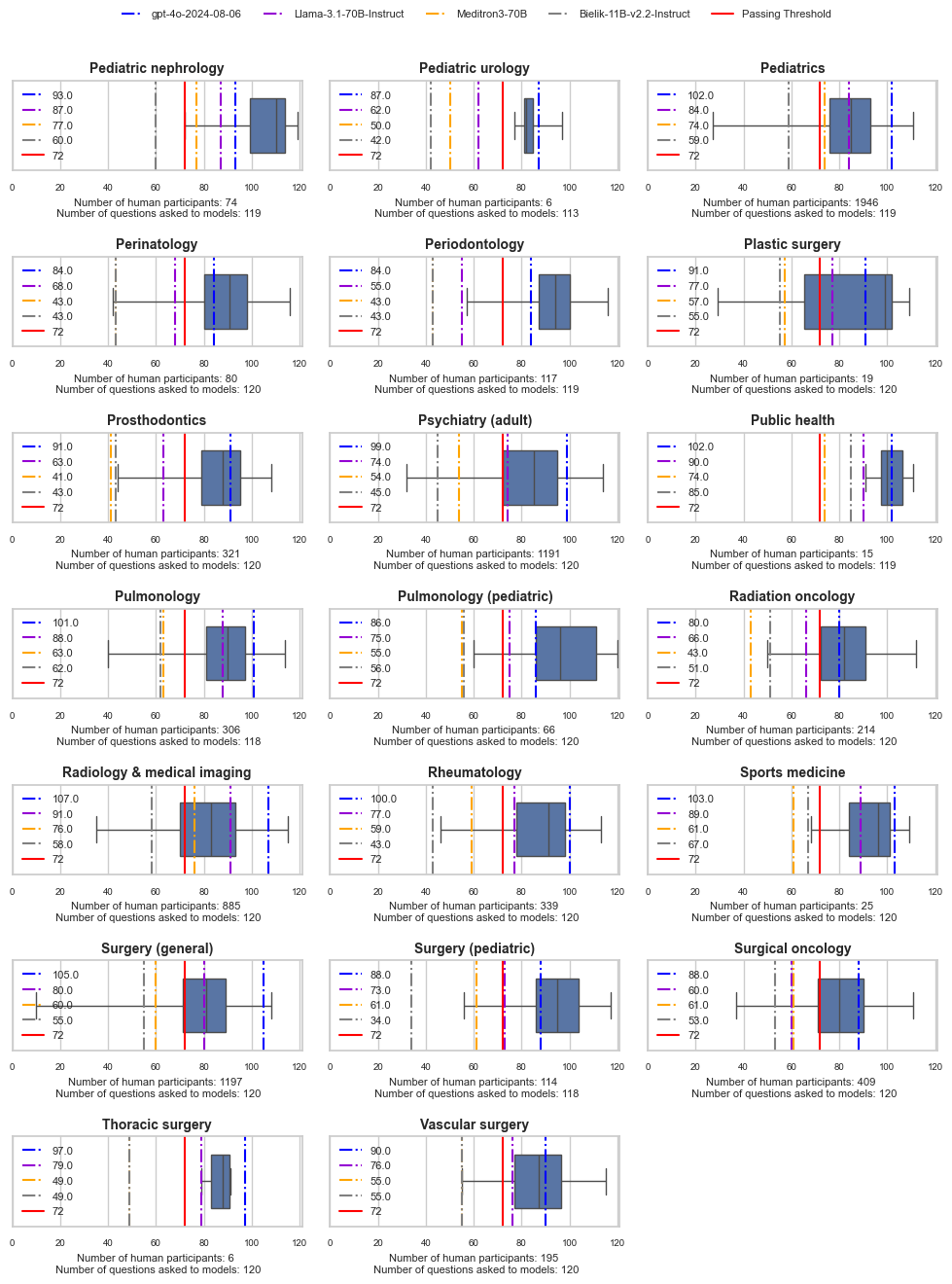}
    \caption{Students performance compared to top-performing LLMs on different specialties on PES exam (part 3/3).}
    \label{fig:pes_human_llm_3}
\end{figure}

\end{document}